\documentclass[ 
twocolumn,
]{ceurart}

\usepackage{listings}
\usepackage{amsmath}
\usepackage{amssymb}
\usepackage{mathtools}
\usepackage{amsthm}

\usepackage{amsmath,amssymb,amsfonts}
\usepackage{algorithmic}
\usepackage{graphicx}
\usepackage{textcomp}
\usepackage{xcolor}
\usepackage[utf8]{inputenc} 
\usepackage{bbm}
\usepackage{nicematrix}
\usepackage{hyperref}
\usepackage{arydshln} 
\usepackage{pifont} 
\usepackage{multirow} 
\usepackage{booktabs} 
\usepackage{diagbox} 
\usepackage{array}    
\usepackage{subcaption}

\usepackage{arydshln}    

\newcommand{\cmark}{\ding{51}} 
\newcommand{\xmark}{\ding{55}} 
\begin{document}

\copyrightyear{2022}
\copyrightclause{Copyright for this paper by its authors.
  Use permitted under Creative Commons License Attribution 4.0
  International (CC BY 4.0).}

\conference{Preprint version}

\title{Deep Joint Distribution Optimal Transport for\\Universal Domain Adaptation on Time Series}


\author[1]{Romain Mussard}[%
email=romain.mussard@univ-rouen.fr,
]
\cormark[1]
\address[1]{Univ Rouen Normandie, INSA Rouen Normandie, Normandie Univ, LITIS UR 4108}

\author[1]{Fannia Pacheco}[]

\author[1]{Maxime Berar}[]

\author[1]{Gilles Gasso}[]

\author[1]{Paul Honeine}[]

\cortext[1]{Corresponding author.}

\begin{abstract}
\normalsize
\noindent Universal Domain Adaptation (UniDA) aims to transfer knowledge from a labeled source domain to an unlabeled target domain, even when their classes are not fully shared. Few dedicated UniDA methods exist for Time Series (TS), which remains a challenging case. In general, UniDA approaches align common class samples and detect unknown target samples from emerging classes. Such detection often results from thresholding  a discriminability metric. The threshold value is typically either a fine-tuned hyperparameter or a fixed value, which limits the ability of the model to adapt to new data. Furthermore, discriminability metrics exhibit overconfidence for unknown samples, leading to misclassifications. 
This paper introduces UniJDOT, an optimal-transport-based method that accounts for the unknown target samples in the transport cost. Our method also proposes a joint decision space to improve the discriminability of the detection module. In addition, we use an auto-thresholding algorithm to reduce the dependence on fixed or fine-tuned thresholds. Finally, we rely on a Fourier transform-based layer inspired by the Fourier Neural Operator for better TS representation. Experiments on TS benchmarks demonstrate the discriminability, robustness, and state-of-the-art performance of UniJDOT.
\end{abstract}


\maketitle

\section{Introduction}

\begin{figure*}[ht]
    \centering
    \includegraphics[width=\linewidth]{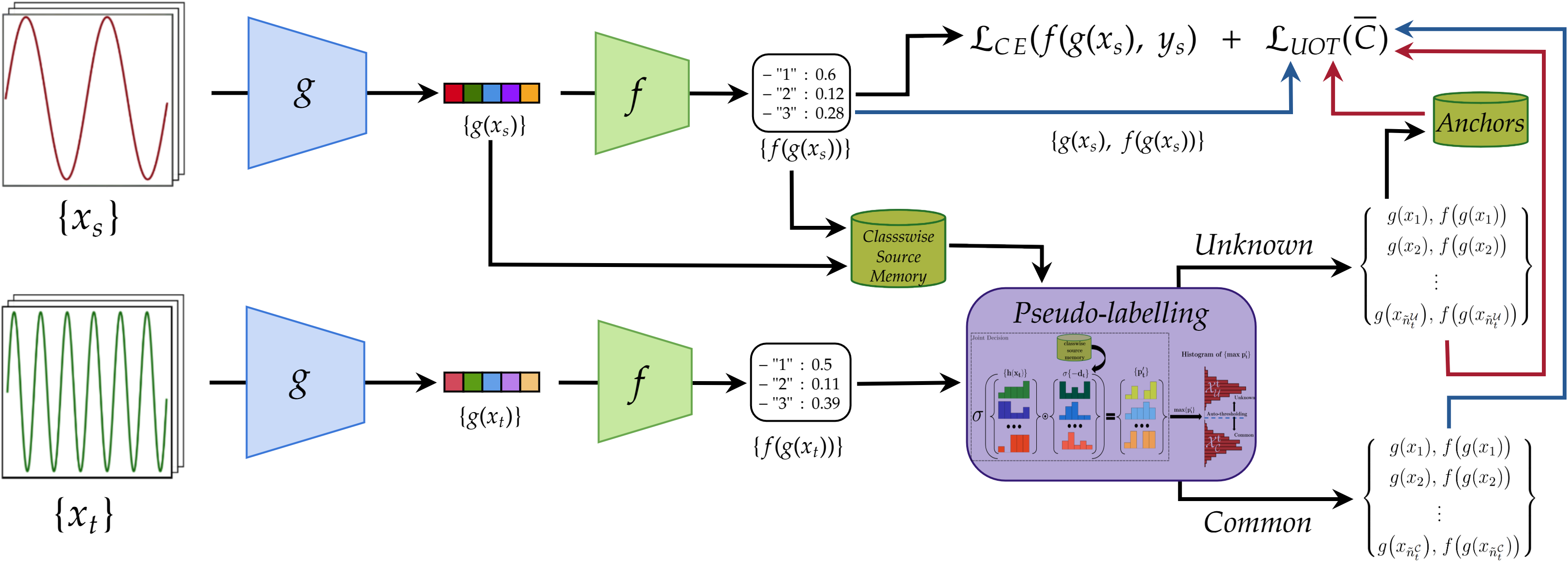}
    \caption{\textbf{Overview of the proposed method:} The source samples $x_s$ and the target samples $x_t$ are processed by the feature extractor $g$ and the classifier $f$, resulting in a feature space representation $g(x_s)$ (resp. $g(x_t)$) and logits $f(g(x_s))$ (resp. $f(g(x_t))$).
    $g(x_s)$ are stored in a classwise memory, while some of the $g(x_t)$ serve as anchors in the alignment process after the pseudo-labelling step.
    The pseudo-labelling step relies on the target logits $f(g(x_t))$ mitigated by a classwise distance of an entire batch to automatically label the target samples as common or unknown. 
    The network is classically trained using a cross-entropy loss on the source samples and an alignment loss between the source and target samples.
    This alignment loss relies on optimal transport and takes into account the pseudo-labelling step as follows: common target samples are aligned with source samples (in \textcolor{blue}{blue}), while unknown target samples are aligned with anchors (in \textcolor{red}{red}).  
    }
    \label{fig:flowchart}
\end{figure*}

Deep learning models have enabled significant improvements in Time Series (TS) classification, due to their powerful representational capabilities \cite{TSsurvey}. However, most models suffer from a lack of generalization and can hardly be transferred from their training domain (the so-called source domain) to an unfamiliar, external domain (the target domain) \cite{ADATIME}. This limitation is mainly caused by distribution shifts, a very common phenomenon in TS \cite{ZHANG202345, GUO2023104998}. To address this problem, Unsupervised Domain Adaptation (UDA) attempts to find a domain-invariant feature space for labeled source samples and unlabeled target samples. 
Recently, several UDA algorithms have been benchmarked for TS tasks \cite{ADATIME}, mainly using image processing architectures. The limited performance of UDA methods on TS datasets compared to their high accuracy on image tasks underscores the need for TS-specific architectures. 
Since \cite{ADATIME}, many approaches have enhanced the performance of TS representation by identifying invariant temporal features specifically tailored for TS \cite{liu2024boosting, ozyurt2023contrastive,lee2024soft, biswas2024, match_and_deform}. For example, frequency-based features extracted from neural operators, such as Fourier Neural Operators (FNO) \cite{fno_li2020fourier}, have improved domain-invariant representations and have been used for UDA on TS \cite{RAINCOAT}.

The UDA framework actually covers a special case where the classes are the same in both domains (despite some small drift between domains) known as the closed-set assumption. Motivated by this limitation, Universal Domain Adaptation (UniDA) was introduced \cite{UDA}. In UniDA, the target domain remains unlabeled, certain classes are shared between the source and target domains, while others are exclusive to either domain. Unlike standard UDA, UniDA helps to discover unknown class samples that are unique to the target domain. This capability is crucial in TS classification, where identifying unseen patterns is often critical \cite{TSsurvey2}.

UniDA encompasses two primary tasks: alignment of common classes across domains and isolation of unknown Out-Of-Distribution (OOD) samples. Several approaches have been developed to address both tasks using different strategies.
The alignment module is usually a pre-existing UDA method that is slightly modified to account for unknown labels. For instance, UAN \cite{UDA} uses an adversarial UDA approach introduced in \cite{ganin2016, tzeng2017}, while DANCE \cite{DANCE} adopts a similarity-based clustering approach inspired by \cite{haeusser2017}. Similarly, PPOT \cite{PPOT} and UniOT \cite{UniOT} align source and target common samples using Optimal Transport (OT) \cite{villani2009optimal}, as proposed in \cite{deepJDOT, JUMBOT}. Alternatively, UniAM \cite{UniAM} introduces a novel alignment framework leveraging sparse data representation and Vision Transformer (ViT) architectures. 

On the other hand, the cited approaches rely on OOD detection methods that are highly threshold-dependent. These methods aim to separate common from unknown samples by thresholding discriminability metrics such as entropy \cite{DANCE, UDA}, OT masses \cite{PPOT, UniOT}, reconstruction error \cite{UniAM}, or combinations of metrics \cite{CMU}. The threshold value is typically either a fine-tuned hyperparameter \cite{UDA, DANCE, PPOT, UniAM} or a fixed value \cite{UniOT} determined from a dataset. However, as shown in experiments in Section \ref{IV}, the optimal threshold often varies between datasets and across tasks. Nevertheless, most previous approaches rely on robust state-of-the-art alignment modules to address the inefficiencies in OOD detection caused by rigid threshold definitions. These approaches fail to consider that inadequate OOD detection can compromise the performance of the alignment module. In consequence, these observations highlight the need for a dynamic threshold selection to better adapt UniDA methods to different datasets.

Few approaches attempt to address the dynamic nature of the threshold value. For example, TNT \cite{TNT} computes a task-specific threshold, but this computation is based solely on the source dataset. Domain shifts may render such computations ineffective on the target dataset. Moreover, OVANet \cite{OVANET} employs a One-vs-All strategy that eliminates the need for a threshold to distinguish common from unknown samples \cite{padhy2020revisiting}. In addition to the challenge of defining a threshold, CMU \cite{CMU} states that relying on a single discriminability metric for OOD detection (such as the entropy used in UAN) is not sufficiently accurate and proposes a combination of metrics (entropy, consistency, and confidence). The lack of discriminability of a single metric often arises from the overconfidence exhibited by deep learning-based models \cite{lakshminarayanan2017simple}, which limits their ability to generate uncertainty spaces for unknown samples

Finally, the previously described UniDA methods were originally designed for computer vision tasks.  To the best of our knowledge, the only UniDA approach tailored for TS is RAINCOAT \cite{RAINCOAT}, which introduces a modification of the FNO layer.
In this method, an OT alignment is performed based on the variation of the target samples' latent representations in a deep reconstruction scheme.  This suggests that the presence of private class samples will modify the representation space after some optimization under a reconstruction loss.
Such variation is then measured by a statistical test for bimodality on the target batch samples. The underlying RAINCOAT assumption is that: pseudo-labelling common and unknown classes is possible at the batch level.  This approach is computationally expensive due to its reliance on this internal deep learning optimization step. However, some elements can be retained to approach UniDA on TS: the FNO architecture and its pseudo-labelling assumption. Using FNO allows us to build a UniDA approach tailored for TS, as in \cite{RAINCOAT}.

In this paper, we introduce \textbf{UniJDOT}, a novel OT-based UniDA method for times series classification. This method can be viewed as an extension of DeepJDOT \cite{deepJDOT} to UniDA, with a focus on TS.
The main novelties associated with our proposed method are:

\begin{itemize}
\item A joint decision space that mitigates classifier outputs using distance-based probability vectors over the feature space. This is designed to cope with the lack of discriminativity described by CMU \cite{CMU}.
    
\item The use of a binary auto-thresholding approach on the target batch samples to pseudo-label them into two classes: common and unknown. This reduces the dependence on hyperparameters and allows for a more robust training.

\item Finally, we introduce an OT-cost rewriting scheme that jointly oversees the alignment of common samples and the isolation of unknown samples. 
\end{itemize}

An overview of the proposed method is shown in \figurename~\ref{fig:flowchart}, and can be summarized as follows. Along with convolutional neural networks (CNNs), a layer inspired by the FNO is used to capture time-frequency features \cite{RAINCOAT}. A pseudo-labelling block separates unknown and common target samples. Finally, common target classes are aligned with their source counterparts with OT, while unknown target samples are effectively isolated in a decision space. 
The experiments highlight the significance of each proposed module and demonstrate the superior performance of UniJDOT compared to state-of-the-art methods applied to TS.

The remainder of this paper is organized as follows: Section~\ref{II} formalizes the discrete OT, FNO-based architecture, and domain adaptation with DeepJDOT. Section~\ref{III} describes UniJDOT. Section~\ref{IV} presents experimental evaluations on TS benchmarks. Finally, Section~\ref{V} concludes the paper. Our code is available at \url{https://github.com/RomainMsrd/UniJDOT}.

\section{Background} \label{II}

This section provides an overview of key concepts relevant to our approach, including Fourier Neural Operators, Discrete Optimal Transport, and Deep Joint Distribution Optimal Transport. Additionally, we formally define Universal Domain Adaptation.

\subsection{Fourier Neural Operator}
The Fourier Neural Operator (FNO) was introduced in \cite{fno_li2020fourier} with a focus on approximating the solution operators of partial differential equations. The FNO relies on multiple Fourier layers, which consist of projecting the output of a neural network into the Fourier space by applying a Fast Fourier Transform. Following this step, a linear transformation is applied to the lower Fourier modes while the higher modes are filtered. Finally, an Inverse Fast Fourier Transform is applied to project back the data into the original space. The original Fourier layer is slightly modified in \cite{RAINCOAT}, by using a cosine smoothing function to prevent alignment over noisy frequency features and extracting polar coordinates of the frequency coefficients instead of performing an inverse Fourier transform.
In an abuse of notation, this modified Fourier layer will be referred to as FNO in this paper.

\subsection{Discrete Optimal Transport}

Let $\boldsymbol{\alpha}$ and $\boldsymbol{\beta}$ be two empirical probability distributions of supports $\mathcal{X}_s = \{x_i^s \in \mathbb{R}^d\}_{i=1}^{n_s}$ and $\mathcal{X}_t = \{x_i^t \in \mathbb{R}^d\}_{i=1}^{n_t}$ such that $\boldsymbol{\alpha} = \sum_{i=1}^{n_s} \mathbf{a}_i \delta_{x_i^s}$ and $\boldsymbol{\beta} = \sum_{i=1}^{n_t} \mathbf{b}_i \delta_{x_i^t}$ with $\mathbf{a} \in \mathbb{R}^{+}_{n_s}$, $\mathbf{b} \in \mathbb{R}^{+}_{n_t}$ and $\delta_{x_i}$ the Dirac function at sample $x_i$.
OT tackles the problem of computing a transport plan $\gamma$ between  $\boldsymbol{\alpha}$ and $\boldsymbol{\beta}$ assigning source samples to target samples.  With $C_{ij}$ the cost associated with moving $x_i^s$ toward $x_j^t$, it is formalized as: 
\begin{equation}
\textbf{OT}(\mathbf{a}, \mathbf{b}, \mathbf{C}) = \underset{\gamma \in \Pi(\mathbf{a},\mathbf{b})}{\min} \langle \mathbf{C}, \gamma \rangle_F \text{,}
\label{eq:kanto}
\end{equation}
where $\Pi(\mathbf{a},\mathbf{b}) = \{ \gamma \in (\mathbb{R}^{+})^{n_s \times n_t} | \gamma \mathbbm{1}_{n_t} = \mathbf{a}, \gamma ^{\top}\mathbbm{1}_{n_s} = \mathbf{b} \}$ and $\mathbf{C} \in \mathbb{R}^{n_s \times n_t}$. By adding an entropic regularization to \eqref{eq:kanto}, this problem becomes tractable on large amounts of data using the Sinkhorn algorithm \cite{sinkhorn}, thereby increasing the use of OT in data science \cite{OT_book}.

Unbalanced Optimal Transport (UOT) \cite{UOT_gilles, UOT_chizat} proposes a relaxation of the mass preservation enforced by the set $\Pi(\mathbf{a},\mathbf{b})$: 
\begin{equation}
\begin{split}
\textbf{UOT}(\mathbf{a}, \mathbf{b}, \mathbf{C}) = \underset{\gamma \geq 0}{\min} \langle \mathbf{C},  \gamma \rangle  & + \tau_1 \text{KL}(\gamma\mathbbm{1}_{n_t}, \mathbf{a}) \\ 
&+ \tau_2 \text{KL}(\gamma^\top \mathbbm{1}_{n_s}, \mathbf{b}),
\end{split}
\end{equation}
where $\tau_1$ and $\tau_2$ are penalization coefficients and KL is the Kullback-Leibler divergence. 
This formulation is deemed more robust to distribution shift and unbalanced mini-batch \cite{JUMBOT}. In our case, the masses $\mathbf{a}$ and $\mathbf{b}$ are each uniform, since there is no reason to assign different weights to individual samples.
Without loss of generality, we will denote $\textbf{UOT}(\mathbf{a}, \mathbf{b}, \mathbf{C})$ by $\mathcal{L}_\text{UOT}(\mathbf{C})$ and $\textbf{OT}(\mathbf{a}, \mathbf{b}, \mathbf{C})$ by $\mathcal{L}_\text{OT}(\mathbf{C})$.

\subsection{Domain Adaptation with DeepJDOT}
\label{section:DeepJDOT}

Let $g$ be a feature extractor that maps the input space into a feature space, and $f$ be a classifier that maps the feature space to the label space. Let $h(x) = f(g(x))$.
DeepJDOT \cite{deepJDOT} proposes to minimize the joint discrepancy between the distributions of each domain seen in both the feature and label spaces while optimizing the classification accuracy on the target domain. This can be done by defining the following transportation cost for each source sample $x_i^s$ and target sample $x_j^t$:
\begin{equation}
{C}_{ij} = \mu \lVert g(x_i^s) - g(x_j^t)\rVert_2^2 + \lVert y_i^s-h(x_j^t)\rVert_2^2,
\label{eq:3}
\end{equation}
where $\mu$ is a tradeoff parameter between the feature and label space distances, and $y^s_i$ is the label of the source sample $i$. The training loss 
$\mathcal{L}$ is a $\lambda$-weighted sum of the cross-entropy loss ($\mathcal{L}_{ce}$) over the source samples combined with $\mathcal{L}_{OT}$: 
\begin{equation} 
    \mathcal{L} = \lambda \mathcal{L}_{ce} + (1-\lambda) \mathcal{L}_\text{OT}(\mathbf{C}).
    \label{eq:4}
\end{equation}
\noindent
More precisely, this leads to the minimization problem:
\begin{equation} 
    \underset{\gamma, f, g}{\text{min}}~ \frac{\lambda}{n_s} \sum_i \mathcal{L}_{ce}(y_i^s, f(g(x^s_i)) + (1-\lambda) \sum_{i,j} \gamma_{i,j} {C}_{ij}.
    \label{eq:5}
\end{equation}
The optimization scheme follows an alternating approach: first set $\gamma$, then set $f$ and $g$.
Setting $f$ and $g$ reduces the problem to an OT minimization \eqref{eq:kanto}, while setting $\gamma$ makes optimizing $f$ and $g$ a standard deep learning task. Each problem is solved alternatively. DeepJDOT is a UDA method that assumes that the classes in both the source and target domains are the same.
The alignment of the domains results in a small discrepancy between the source and target distributions at the end of the training process.

\subsection{Universal Domain Adaptation}

UniDA extends the scope of UDA by addressing a more general and challenging setting \cite{UDA}. In UniDA, we are given a labeled source domain,  
$\mathcal{D}^s = \{(x_i^s, y_i^s)\}_{i=1}^{n_s},$ which follows a joint distribution \( \mathcal{P}^s(x^s, y^s) \), and an unlabeled target domain,  
$\mathcal{D}^t = \{(x_i^t)\}_{i=1}^{n_t},$ which follows another joint distribution \( \mathcal{P}^t(x^t, y^t) \). As in UDA, it is assumed that the source and target distributions are not identical, i.e., \( \mathcal{P}^s \neq \mathcal{P}^t \).
UniDA operates under the flexible assumption that the label sets may differ, i.e., \(\mathcal{Y}^s \neq \mathcal{Y}^t\). 
To capture this distinction, the label sets are divided into three subsets \cite{UDA}:  

\begin{itemize}  
    \item The \textbf{common label set} (\(\mathcal{Y}^C = \mathcal{Y}^s \cap \mathcal{Y}^t\)), which contains labels shared by the source and target domains.  
    \item The \textbf{private source label set} (\(\overline{\mathcal{Y}^s} = \mathcal{Y}^s \backslash \mathcal{Y}^t\)), which consists of labels that exist only in the source domain.  
    \item The \textbf{private target label set} (\(\overline{\mathcal{Y}^t} = \mathcal{Y}^t \backslash \mathcal{Y}^s\)), consisting of labels unseen in the source domain and are only present in the target domain.  
\end{itemize}  
In this paper, we propose a novel methodology designed to solve UniDA problems.

\section{Universal Deep-Joint Distribution Optimal Transport} \label{III}

\underline{Uni}versal DA with Deep\underline{JDOT} (UniJDOT) extends DeepJDOT by rewriting the alignment cost to take into account the unknown target samples. 
Three key steps are performed: i) Pseudo-labeling, where unlabeled target data are divided into unknown and common, ii) Anchor determination, to provide a correspondence in the source domain for unknown target samples and iii) Domain alignment through UOT.

\begin{figure}[t]
    \centering
    \includegraphics[width=\linewidth]{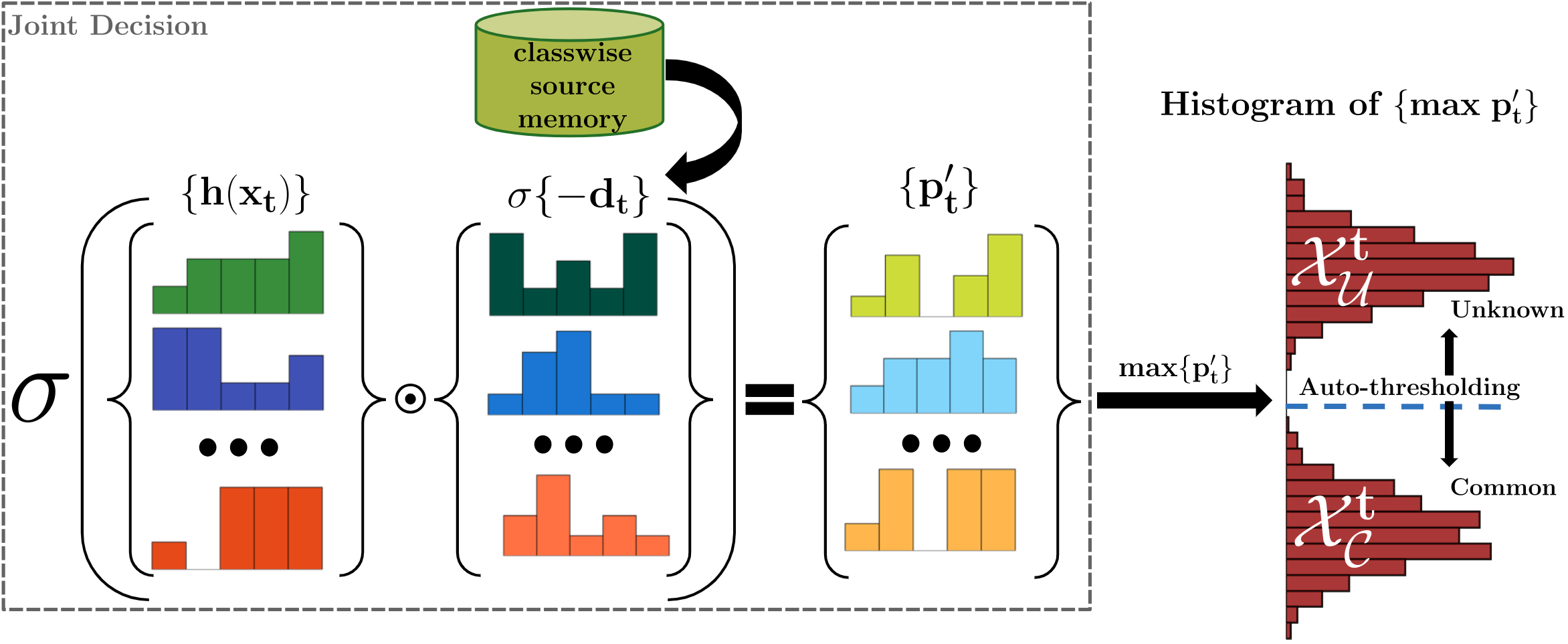}
    \caption{\textbf{Pseudo-labelling:} 
    Each logit $h(x_t)$ is multiplied by a distance-based probability vector $\sigma(-d_t)$ computed using a classwise memory,  resulting in the batch $\left\{ p'_t \right\}$. Then, a binary auto-thresholding is applied on the distribution of $\left\{\max ~ p'_t \right\}$ labelling the target samples of the batch.}
    \label{fig:flowchart_pseudo}
\end{figure}

\subsection{Pseudo-labelling}

Given a pretrained network $h$ on the source domain, our goal is to discriminate between unknown and common target data based on the classifier's outputs. However, the classifier often exhibits overconfidence when dealing with unknown samples \cite{lakshminarayanan2017simple}. To mitigate this, we propose a merged approach that regularizes the classifier's predictions by integrating distance-based probabilities within the feature space, while leveraging all target samples in a given batch. 
We assume that at the batch level, samples of both common and unknown classes exist. By doing so, we ensure that a relatively large distance in the feature space is associated with low model confidence.

As illustrated in \figurename~\ref{fig:flowchart_pseudo}, each logit $h(x_t)$ of the batch is adjusted by a distance-based probability vector, resulting in a collection of probability vectors $p_t'$. The maximum value of each probability vector in $p_t'$ is then assembled into a histogram. Based on this histogram, an automatic thresholding approach separates the samples of the target batch into common and unknown samples.

More precisely, for a given sample $x_t$, we condition the classifier outputs $h(x_t)$ through element-wise multiplication with distance-based probabilities. The softmax function $\sigma$ is then applied to obtain a probability vector:
$$
p_t' = \sigma \left(h(x^t) \sigma(-d_t) \right),
$$
\noindent where $d_t$ is a vector containing the distances between a target sample $x^t$ and its nearest neighbor in $\mathcal{X}_s^k$, which is the set of source samples of class $k$, for all $k \in \{1,...,K\}$ classes:
$$
d_t = \left( \underset{x^s \in \mathcal{X}_s^1}{\text{min}} \, d(x^t, x^s), \cdots, \underset{x^s \in \mathcal{X}_s^K}{\text{min}} \, d(x^t, x^s)\right).
$$

\noindent Calculating $d_t$ requires computing the pairwise distance between the target sample and a collection of source samples. Therefore, it is critical to have at least one example of each class in the collection of source samples to ensure that each value of \(d_t\) is well-defined. To address this, we introduce a memory mechanism (see Fig. \ref{fig:flowchart_pseudo}) that holds \(N_c\) source samples for each class in the source domain. This memory is initialized before training and ensures that there are enough source samples per class to compute the distance vector \(d_t\) appropriately. 
The memory is continuously updated as new source batches are fed into the model during training.

\begin{figure}[t]
    \centering
    
    \begin{subfigure}{0.45\linewidth}
        \centering
        \includegraphics[width=\linewidth]{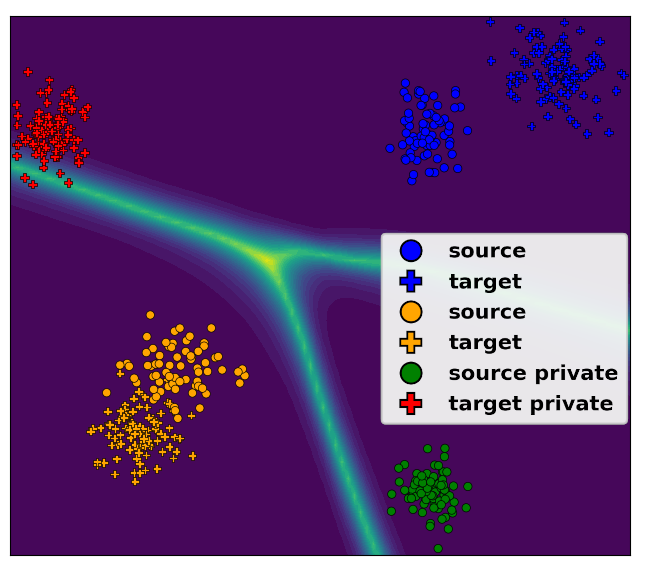}
        \caption{Softmax Decision}
        \label{fig:no_joint}
    \end{subfigure}
    \begin{subfigure}{0.52\linewidth}
        \centering
        \includegraphics[width=\linewidth]{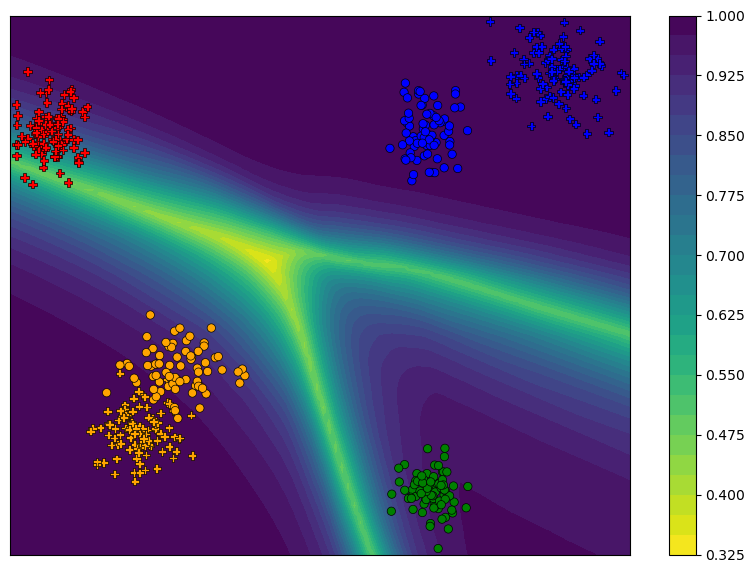}
        \caption{Joint Decision}
        \label{fig:joint}
    \end{subfigure}
    
    \caption{\textbf{Illustration of the decision space on a 2D toy dataset:} The color represents the confidence level, determined by either a) the maximum value of $\sigma(h(x))$ or b) the maximum value of $p'_t$. A lighter color corresponds to less confidence.}
    \label{fig:decision_plots_combined}
\end{figure}

As shown in Fig. \ref{fig:decision_plots_combined}, the proposed joint decision space increases the indecision margin compared to a decision space that relies solely on the model's output. The smoother confidence distribution around unknown samples, thereby reduces the risk of mislabelling such samples.

For a given batch, the maximum value of each probability vector $p_t'$ is collected into an empirical scalar distribution. Such a distribution represents the prediction confidence of the model. Given our previous assumption, one can find a threshold that maximizes the separation between samples with low and high prediction confidence. Such a problem is well known in image processing, and several approaches have been developed under the so-called binary auto-thresholding (e.g. Otsu \cite{otsu}, Li \cite{li_thr, li_thr_iterative}, Yen \cite{yen_thresh}, and triangle \cite{triangle_method}). 

Finally, the pseudo-labels $\kappa_t$ are obtained using the threshold $\tau$ returned by the implemented auto-thresholding method:
$$
\kappa_t = 
\begin{cases} 
\text{Unknown} & \text{if } \max p_t' < \tau \\
\text{Common} & \text{otherwise}
\end{cases}
$$
This results into two target sets defined at the batch level $\mathcal{X}_{\mathcal{C}}^t$ (common) and $\mathcal{X}_{\mathcal{U}}^t$ (unknown).

\subsection{Anchors}
Only target samples in $\mathcal{X}_{\mathcal{C}}^t$ should be aligned with source samples. Therefore, we need to introduce anchors to align target samples in $\mathcal{X}_{\mathcal{U}}^t$ to avoid misalignment with source samples. Since DeepJDOT relies on joint alignment in the feature space and the decision space, anchors should be defined in both spaces. On the one hand, due to the well-defined structure of the decision space as a simplex, we choose  $r=\frac{1}{|\mathcal{C}|}\mathbbm{1}$ as the only decision space anchor  since it corresponds to the most uncertain class probability vector, representing lowest confidence for any classifier.
On the other hand, to define anchors in the feature space, we propose to use multiple anchors of feature vectors $\{a_l\}_{l=1}^{L}$ computed from unknown target samples. At initialization, 
a K-means algorithm computes the $L$ centroids of all the target sample feature vectors that form the anchors set. For each batch at the training stage, this set is updated using a moving average to reduce the computational complexity. The number $L$ takes into account that unknown samples may belong to multiple clusters in the feature space.

\subsection{Alignment}
For any common target sample $x^t_j \in \mathcal{X}^t_{\mathcal{C}}$, the DeepJDOT cost \eqref{eq:3} is used and provides the block $\mathbf{C}^{\mathcal{C}}$, defined with 
\begin{equation}
C_{ij}^{\mathcal{C}} = \mu \lVert g(x_i^s) - g(x_j^t)\rVert_2^2 + \lVert h(x_i^s)-h(x_j^t)\rVert_2^2,
\end{equation}
while, for any unknown sample $x^t_j \in \mathcal{X}_{U}^t$, a new cost is defined relying on the anchors $a$ and $r$:
\begin{equation}\label{eq:unknown}
C_{kj}^{\mathcal{U}} = \mu \lVert a_k - g(x_j^t)\rVert_2^2 + \lVert r-h(x_j^t)\rVert_2^2.
\end{equation}
Based on these propositions, we need to form a new cost matrix $\Bar{\mathbf{C}}$ that enables both formulations:
\begin{equation}
    \Bar{\mathbf{C}} = 
\begin{bNiceArray}{cc}
\mathbf{C}^{\mathcal{C}} & \xi \mathbbm{1}_{\tilde{n}_s, \tilde{n}_t^{\mathcal{U}}}\\
\xi \mathbbm{1}_{L, \tilde{n}^{\mathcal{C}}_t}   &   \mathbf{C}^{\mathcal{U}} \\
\end{bNiceArray},
\end{equation}
where
\begin{itemize}
\item $\mathbf{C}^{\mathcal{C}} \in \mathbb{R}^{\tilde{n}_s \times \tilde{n}_t^{\mathcal{C}}}$ aligns the $\tilde{n}_t^{\mathcal{C}}$ common target samples with the $\tilde{n}_s$ source samples of the mini-batch, 
\item $\mathbf{C}^{\mathcal{U}} \in \mathbb{R}^{L \times \tilde{n}_t^{\mathcal{U}}}$ is the block assigned to the unknown samples based on \eqref{eq:unknown}, aligning the $\tilde{n}_{t}^{\mathcal{U}}$ unknown target samples of the mini-batch with the anchors, 
\item $\mathbbm{1}_{\tilde{n}_s, \tilde{n}_t^{\mathcal{U}}}$ and $\mathbbm{1}_{L, \tilde{n}^{\mathcal{C}}_t}$ are unit matrices of appropriate sizes,
and $\xi \geq \max \left\{ \max_{i,j} {C}^{\mathcal{C}}_{i,j}, \max_{i,j} {C}^{\mathcal{U}}_{i,j} \right\}$.
\end{itemize}

Finally, the loss of our model is the same as DeepJDOT (see \eqref{eq:4}) except that UOT is used instead of OT with the new cost matrix $\Bar{\mathbf{C}}$:
\begin{equation*} \label{eq:6}
    \mathcal{L} = \lambda \mathcal{L}_{ce} + (1-\lambda) \mathcal{L}_\text{UOT}(\bar{\mathbf{C}}).
\end{equation*}

\subsection{Inference}

The pseudo-labelling module is inherently batch-dependent during training. However, batches as such do not exist during inference. Therefore, the pseudo-labelling assumption does not hold. In order to address this limitation, we propose a method to set the threshold value after training based on the target training data used for adaptation. Specifically, after training, we can retain the last threshold computed by the model and use it during inference. Alternatively, we adopt a more robust approach that uses a larger validation batch and retains the automatically determined threshold.

\section{Experiments} \label{IV}

\subsection{Experimental Settings}

We replicate the standard UniDA experimental setup described in \cite{PPOT, UniOT, OVANET, CMU}, and adapt it to the TS datasets referenced in the UDA benchmarking paper \cite{ADATIME}. The UniDA framework involves artificially creating UniDA tasks by removing labels from the source and target datasets to simulate the emergence of unknown classes. To the best of our knowledge, this is the first time this framework has been applied to TS datasets.
As a result, only a limited number of the 5 TS datasets listed in \cite{ADATIME} are suitable for this framework. For example, WISDM \cite{wisdm} contains very few samples, while FD \cite{boiler} has only three classes available, making them unsuitable for this study. In contrast, HAR \cite{HAR_dataset}, HHAR \cite{HHAR_dataset}, and Sleep-EDF (EDF) \cite{EEG_dataset} were the only datasets deemed appropriate. However, due to the limited number of classes, we designated one class as source-private (present only in the source) and another as target-private (present only in the target) for each dataset.
We provide a detailed description of each considered TS dataset below:
\begin{itemize}
    \item \textbf{HAR} \cite{HAR_dataset}. TS samples were collected from 30 participants at a sampling rate of 50Hz using the smartphone's embedded sensors: a 3-axis accelerometer, a 3-axis gyroscope, and a 3-axis body acceleration sensor. Each participant is considered as one domain. Each domain contains 9 channels, divided into non-overlapping windows of 128 time steps. The classification task is to recognize one of six activities in each segment: walking, walking upstairs, walking downstairs, sitting, standing, or lying down. 

    \item \textbf{HHAR} \cite{HHAR_dataset}. This dataset was collected from 9 participants using the smartphone's embedded 3-axis accelerometers. Similar to the HAR dataset,  these signals are divided into non-overlapping segments of 128 time steps. However, only 3 channels are available. The classification task is to identify one of six activities: biking, sitting, standing, walking, walking upstairs, or walking downstairs. Each participant is considered as one domain.
    
    \item \textbf{Sleep-EDF} \cite{EEG_dataset}. This dataset contains electroencephalography recordings from 20 patients. The classification goal is to determine the sleep stage for each segment, with five possible stages. Each sequence is univariate and spans 3{,}000 time steps. Each patient is considered a domain.
\end{itemize}
We use the preprocessed versions of these datasets proposed by \cite{ADATIME}, which already include a 70\% / 30\% train/test split. 

UniDA performance is measured as in \cite{PPOT, UniOT, OVANET}, using the H-score defined as $(2A_\text{C}A_\text{U})/(A_\text{C}+A_\text{U})$, where $A_\text{C}$ is the accuracy on target common classes and $A_\text{U}$ on target unknown classes. We compare \textbf{UniJDOT} with five UniDA  state-of-the-art baselines: \textbf{UAN}, \textbf{OVANet}, \textbf{DANCE}, \textbf{PPOT}, and \textbf{UniOT}.  It is important to mention that RAINCOAT's 
reproducibility issues with the UniDA task prevented us from including it in the experiments. These problems were also reported by several researchers\footnote{See: \href{https://github.com/mims-harvard/Raincoat/issues/10}{https://github.com/mims-harvard/Raincoat/issues/10}}.

Since our method uses a CNN+FNO feature extractor, we tested both CNN-only and CNN+FNO for each baseline and selected the best-performing architecture. All models were pretrained on the source data for 20 epochs. For each model, we performed a thorough hyperparameter search using a Bayesian hyperparameter search over 200 hyperparameter combinations.  The search was conducted over 5 validation scenarios from HAR, different from the scenarios used to train and test our model in Section \ref{IV:B}. The same set of hyperparameters was reused for HHAR and EDF.

\begin{table}[t]
\centering
\caption{  H-scores (\%) for HAR}
\resizebox{\linewidth}{!}{
\begin{tabular}{ l c c c c c c}
\toprule
\textbf{Scenario} & \textbf{UAN} & \textbf{OVANet} & \textbf{PPOT}$^\star$ & \textbf{DANCE} & \textbf{UniOT}$^\star$ & \textbf{UniJDOT}$^\star$\\
\midrule 
$12 \rightarrow 16$ & \textbf{57 $\pm$ 06} & 19 $\pm$ 19 & \underline{53 $\pm$ 10} & 34 $\pm$ 25 & 30 $\pm$ 15 & 50 $\pm$ 13 \\
$13 \rightarrow 3$ & \underline{72 $\pm$ 04} & 38 $\pm$ 31 & 53 $\pm$ 37 & \textbf{85 $\pm$ 10} & 69 $\pm$ 05 & 69 $\pm$ 11 \\
 $15 \rightarrow 21$ & \underline{77 $\pm$ 19} & 30 $\pm$ 32 & 52 $\pm$ 39 & \textbf{91 $\pm$ 06} & \underline{77 $\pm$ 02} & 75 $\pm$ 06 \\
 $17 \rightarrow 29$ & 69 $\pm$ 07 & 17 $\pm$ 28 & \textbf{77 $\pm$ 11} & 71 $\pm$ 25 & 71 $\pm$ 03 & \underline{73 $\pm$ 05} \\
 $1 \rightarrow 14$ & \textbf{80 $\pm$ 04} & 06 $\pm$ 10 & 48 $\pm$ 25 & 07 $\pm$ 12 & \underline{64 $\pm$ 21} & 44 $\pm$ 33 \\
 $22 \rightarrow 4$ & \underline{74 $\pm$ 06} & 48 $\pm$ 25 & 61 $\pm$ 34 & \textbf{82 $\pm$ 02} & 67 $\pm$ 06 & 71 $\pm$ 08 \\
 $24 \rightarrow 8$ & 41 $\pm$ 12 & 09 $\pm$ 17 & \textbf{59 $\pm$ 08} & \underline{58 $\pm$ 11} & 55 $\pm$ 11 & 47 $\pm$ 20 \\
 $30 \rightarrow 20$ & 37 $\pm$ 11 & 20 $\pm$ 18 & \underline{49 $\pm$ 17} & 19 $\pm$ 27 & 34 $\pm$ 14 & \textbf{50 $\pm$ 07} \\
 $6 \rightarrow 23$ & 24 $\pm$ 10 & 29 $\pm$ 29 & \textbf{76 $\pm$ 07} & 08 $\pm$ 26 & 53 $\pm$ 14 & \underline{70 $\pm$ 05} \\
 $9 \rightarrow 18$ & \textbf{69 $\pm$ 09} & 42 $\pm$ 28 & 57 $\pm$ 08 & 53 $\pm$ 13 & 49 $\pm$ 07 & \underline{66 $\pm$ 08} \\
\hdashline
 mean & \underline{60} & 26 & 59 & 51 & 57 & \textbf{62} \\

\bottomrule
\multicolumn{4}{l}{$^\star$Models trained with CNN+FNO}\\
\end{tabular}
}
\label{tab:hscore_HAR}
\end{table}

\begin{table}[t]
\centering
\caption{  H-scores (\%) for HHAR}
\resizebox{\linewidth}{!}{
\begin{tabular}{ l c c c c c c}
\toprule
\textbf{Scenario} & \textbf{UAN} & \textbf{OVANet} & \textbf{PPOT}$^\star$ & \textbf{DANCE} & \textbf{UniOT}$^\star$ & \textbf{UniJDOT}$^\star$\\
\midrule 
 $0 \rightarrow 2$ & \textbf{47 $\pm$ 17} & 20 $\pm$ 14 & 01 $\pm$ 01 & 20 $\pm$ 26 & \underline{42 $\pm$ 17} & 40 $\pm$ 19 \\
 $0 \rightarrow 6$ & 43 $\pm$ 10 & 41 $\pm$ 12 & 07 $\pm$ 04 & 46 $\pm$ 28 & \textbf{56 $\pm$ 10} & \underline{48 $\pm$ 11} \\
 $1 \rightarrow 6$ & 44 $\pm$ 15 & 21 $\pm$ 21 & 05 $\pm$ 02 & \underline{64 $\pm$ 15} & 51 $\pm$ 20 & \textbf{66 $\pm$ 11} \\
 $2 \rightarrow 7$ & \textbf{41 $\pm$ 08} & 22 $\pm$ 12 & 10 $\pm$ 03 & 20 $\pm$ 21 & 10 $\pm$ 09 & \underline{28 $\pm$ 06} \\
 $3 \rightarrow 8$ & 58 $\pm$ 20 & 52 $\pm$ 26 & 03 $\pm$ 03 & \underline{69 $\pm$ 21} & 55 $\pm$ 12 & \textbf{71 $\pm$ 12} \\
 $4 \rightarrow 5$ & \underline{48 $\pm$ 16} & 17 $\pm$ 18 & 02 $\pm$ 02 & 02 $\pm$ 02 & 40 $\pm$ 17 & \textbf{57 $\pm$ 18} \\
 $5 \rightarrow 0$ & 16 $\pm$ 09 & 08 $\pm$ 05 & 02 $\pm$ 01 & 00 $\pm$ 01 & \textbf{21 $\pm$ 08} & \underline{17 $\pm$ 14} \\
 $6 \rightarrow 1$ & \underline{77 $\pm$ 14} & 31 $\pm$ 25 & 01 $\pm$ 02 & 62 $\pm$ 42 & 72 $\pm$ 16 & \textbf{79 $\pm$ 13} \\
 $7 \rightarrow 4$ & 45 $\pm$ 10 & 16 $\pm$ 15 & 04 $\pm$ 03 & 06 $\pm$ 04 & \underline{62 $\pm$ 10} & \textbf{64 $\pm$ 19} \\
 $8 \rightarrow 3$ & 42 $\pm$ 31 & 32 $\pm$ 22 & 01 $\pm$ 01 & \textbf{69 $\pm$ 33} & 32 $\pm$ 20 & \underline{56 $\pm$ 18} \\
\hdashline
 mean & \underline{46} & 26 & 03 & 36 & 44 & \textbf{53} \\

\bottomrule
\multicolumn{4}{l}{$^\star$Models trained with CNN+FNO}\end{tabular}
}
\label{tab:hscore_HHAR}
\end{table}

\subsection{Results} \label{IV:B}

Tables \ref{tab:hscore_HAR}, \ref{tab:hscore_HHAR} and \ref{tab:hscore_EEG} respectively report the H-scores of multiple adaptation scenarios for HAR, HHAR and EDF. Best results are reported in bold, and second-best results are underlined. Each scenario was tested over 10 different seeds. The error bars correspond to the standard error over these 10 seeds. The  results are obtained using the best feature extractor (CNN or CNN+FNO) found during the hyperparameter search for each model. Note that for OT-based methods, the CNN+FNO architecture demonstrates better results. On average, our method consistently outperforms all baselines for all datasets. In most scenarios, our approach achieves first or second-best results. 

While several methods, such as UniOT, PPOT, and UAN, show promising results on HAR, the performance gap between these methods and ours tends to widen on HHAR and EDF. Notably, the hyperparameters for all models, including ours, were chosen on randomly selected tasks from the HAR dataset. This observation suggests that these models have overfitted to HAR, resulting in reduced performance on other datasets such as HHAR and EDF.

Many UniDA models, including PPOT \cite{PPOT}, UAN \cite{UDA}, and OVANet \cite{OVANET}, rely on manually selected thresholds to distinguish between common and unknown target samples. This reliance on fine-tuned thresholds appears to be particularly detrimental to PPOT's performance on HHAR, as the model struggles to converge effectively. In contrast, our proposed method, which avoids the use of a fixed or tunable threshold, demonstrates greater robustness across datasets. This adaptability allows our model to maintain superior performance even when evaluated on different datasets, highlighting its generalizability and effectiveness in UniDA scenarios. 

\begin{figure*}[t]
    \centering
    
    \begin{subfigure}{0.49\linewidth}
        \centering
        \includegraphics[width=\linewidth]{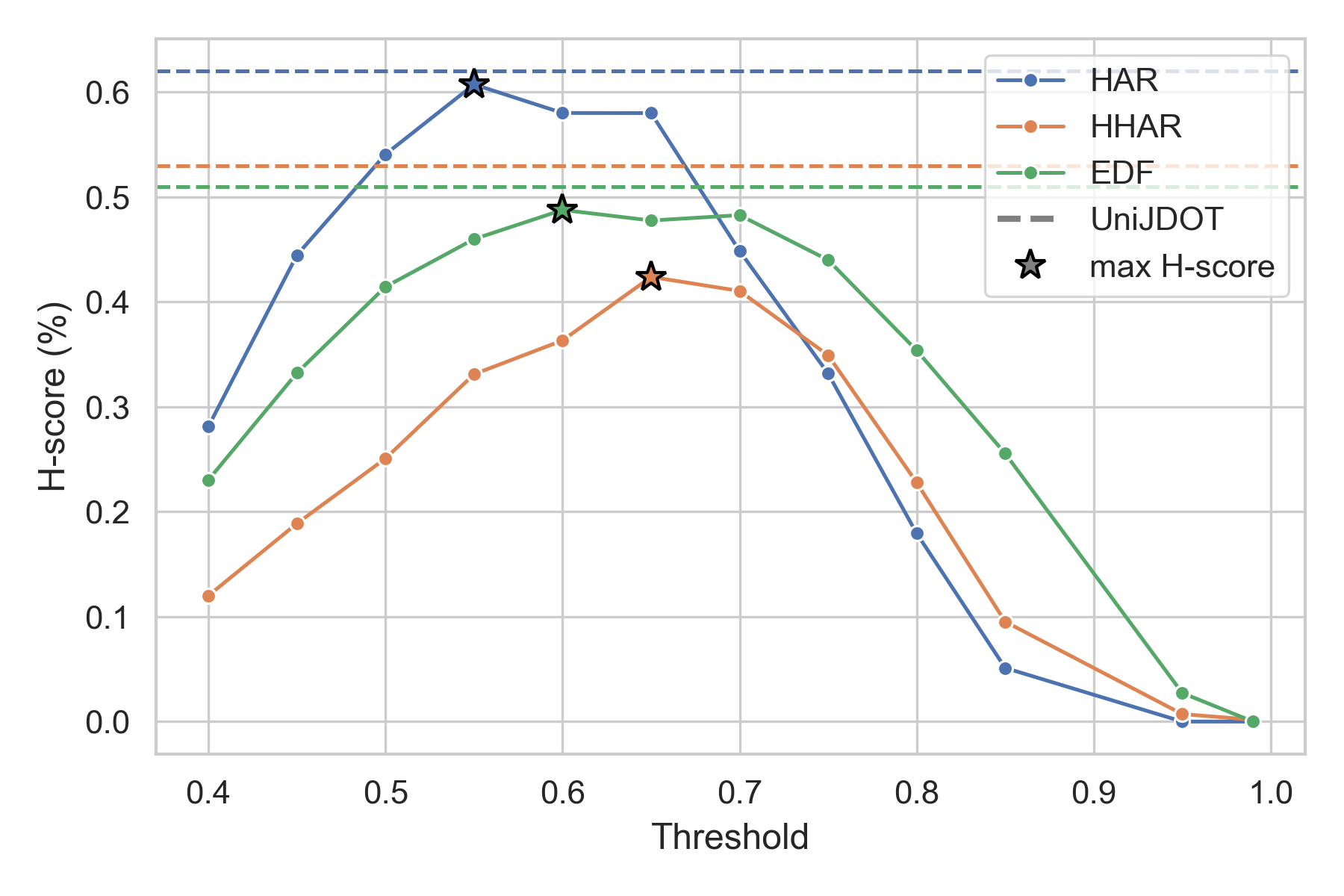}
        \caption{Interdataset threshold sensitivity}
        \label{fig:threshold_plot_inter}
    \end{subfigure}
    \begin{subfigure}{0.49\linewidth}
        \centering
        \includegraphics[width=\linewidth]{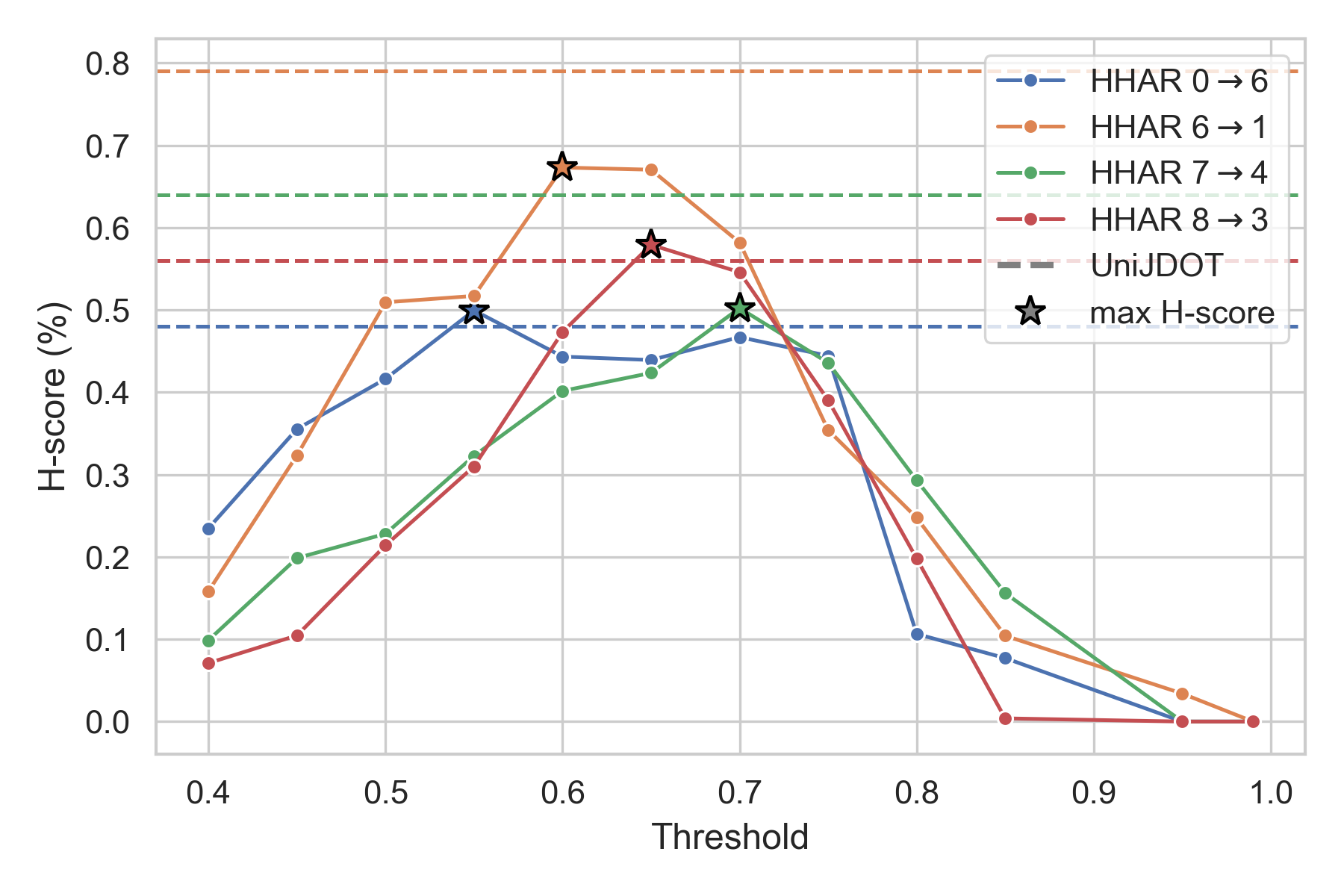}
        \caption{Intradataset threshold sensitivity}
        \label{fig:threshold_plot_intra}
    \end{subfigure}
    
    \caption{\textbf{Threshold sensitivity:} The stars correspond for a) each dataset or b) each scenario. The dot lines show UniJDOT scores when using auto-thresholding. Each color is associated with a dataset.}
    \label{fig:threshold_plots_combined}
\end{figure*}

\begin{table}[t]
\centering
\caption{  H-scores (\%) for EDF}
\resizebox{\linewidth}{!}{
\begin{tabular}{ l c c c c c c}
\toprule
\textbf{Scenario} & \textbf{UAN} & \textbf{OVANet} & \textbf{PPOT}$^\star$ & \textbf{DANCE} & \textbf{UniOT}$^\star$ & \textbf{UniJDOT}$^\star$\\
\midrule 
 $0 \rightarrow 11$ & \textbf{37 $\pm$ 13} & 28 $\pm$ 10 & 05 $\pm$ 09 & 24 $\pm$ 18 & 18 $\pm$ 14 & \underline{35 $\pm$ 13} \\
 $12 \rightarrow 5$ & 51 $\pm$ 10 & 32 $\pm$ 19 & 27 $\pm$ 09 & 54 $\pm$ 13 & \underline{60 $\pm$ 08} & \textbf{65 $\pm$ 05} \\
 $13 \rightarrow 17$ & 32 $\pm$ 06 & 31 $\pm$ 12 & 26 $\pm$ 07 & \underline{50 $\pm$ 15} & 39 $\pm$ 15 & \textbf{54 $\pm$ 11} \\
 $16 \rightarrow 1$ & \underline{45 $\pm$ 05} & 40 $\pm$ 14 & 31 $\pm$ 07 & 25 $\pm$ 12 & 37 $\pm$ 03 & \textbf{50 $\pm$ 05} \\
 $18 \rightarrow 12$ & \underline{31 $\pm$ 04} & 28 $\pm$ 14 & 18 $\pm$ 10 & 20 $\pm$ 07 & 27 $\pm$ 04 & \textbf{33 $\pm$ 04} \\
 $3 \rightarrow 19$ & 37 $\pm$ 03 & \textbf{45 $\pm$ 16} & 23 $\pm$ 06 & 39 $\pm$ 18 & 38 $\pm$ 05 & \underline{42 $\pm$ 10} \\
 $5 \rightarrow 15$ & 36 $\pm$ 10 & 53 $\pm$ 12 & 16 $\pm$ 06 & 42 $\pm$ 27 & \textbf{66 $\pm$ 03} & \underline{61 $\pm$ 04} \\
 $6 \rightarrow 2$ & \textbf{55 $\pm$ 02} & 36 $\pm$ 10 & 30 $\pm$ 11 & 25 $\pm$ 06 & 33 $\pm$ 04 & \underline{42 $\pm$ 04} \\
 $7 \rightarrow 18$ & 53 $\pm$ 02 & 47 $\pm$ 17 & 36 $\pm$ 07 & 31 $\pm$ 11 & \underline{55 $\pm$ 05} & \textbf{56 $\pm$ 02} \\
 $9 \rightarrow 14$ & 43 $\pm$ 04 & 53 $\pm$ 21 & 28 $\pm$ 06 & 62 $\pm$ 16 & \underline{64 $\pm$ 06} & \textbf{70 $\pm$ 05} \\
\hdashline
 mean & 42 & 39 & 24 & 37 & \underline{44} & \textbf{51} \\

\bottomrule
\multicolumn{4}{l}{$^\star$Models trained with CNN+FNO}\end{tabular}
}
\label{tab:hscore_EEG}
\end{table}

\subsubsection{Sensitivity to the threshold}

Several fixed thresholds were tested on UniJDOT to assess the impact of the auto-thresholding component. The H-score for each threshold and dataset is shown in Fig. \ref{fig:threshold_plots_combined}. Additionally, the H-score obtained by UniJDOT is included as a reference for comparison. Fig. \ref{fig:threshold_plot_inter} shows that manually identifying a fixed threshold with the maximum H-score can be challenging, as its value varies from one dataset to another. Similarly, Fig. \ref{fig:threshold_plot_intra} shows that the best threshold also varies greatly from one scenario to another, further complicating the selection process. This implies that a threshold should ideally be set at the scenario level, which is not feasible under the unsupervised assumption of the UniDA framework. Moreover, since the best threshold typically lies within a narrow interval, a hyperparameter search may fail to identify it, as small variations can significantly affect performance. Consequently, the high performance of our method comes from its ability to dynamically compute a threshold for each dataset and scenario at the batch level. This ensures robustness and generalizability in UniDA tasks.

Finally, Table \ref{tab:auto_thresholding_methods} compares several well-known auto-thresholding techniques for image binarization—specifically, Yen \cite{yen_thresh}, Otsu\cite{otsu}, Triangle \cite{triangle_method} and Li \cite{li_thr, li_thr_iterative}. These binarization methods implicitly rely on a hidden hyperparameter: the number of histogram bins. However, this choice is not critical, as the number of bins can be over-parameterized beyond the batch size to achieve finer granularity without negatively impacting performance. Table \ref{tab:auto_thresholding_methods} shows that Yen consistently outperforms the others, indicating that this method is robust across all evaluated datasets compared to the alternatives.
Note that Otsu's is systematically second and Li's third, illustrating the stability and consistency of automated thresholding techniques in the UniDA context.

\begin{table}[t]
\centering
\caption{Comparison of auto-thresholding methods (H-scores)}
\renewcommand{\arraystretch}{1.25} 
\setlength{\tabcolsep}{10pt} 
\begin{tabular}{@{}c@{\hskip 10pt}c@{\hskip 10pt}c@{\hskip 10pt}c@{\hskip 10pt}c@{}}
\toprule
\multirow{2}{*}{\textbf{Datasets}} & \multicolumn{4}{c@{}}{\textbf{Auto-thresholding Methods}} \\ \cmidrule(lr){2-5}
                                   & \textbf{Yen} & \textbf{Otsu} & \textbf{Triangle} & \textbf{Li} \\ \midrule
\textbf{HAR}                       & \textbf{62}       & \underline{55}        & 41            & 50      \\ 
\textbf{HHAR}                       & \textbf{53}       & \underline{40}        & 34            & 38      \\ 
\textbf{EDF}                        & \textbf{51}       & \underline{48}       & 30           & 47     \\ \bottomrule
\end{tabular}
\label{tab:auto_thresholding_methods}
\end{table}

\subsubsection{Ablation study}

The ablation study in Table \ref{tab:ablation} evaluates three core components of our approach: the joint decision mechanism, the auto-thresholding, and the FNO layer. To simulate the absence of auto-thresholding, we replaced it with the best fixed threshold for each dataset, determined directly from the test data. Specifically, we chose the threshold corresponding to the peak performance for each dataset, as shown in Fig. \ref{fig:threshold_plot_inter}. In addition, in the absence of the FNO layer, only CNN layers were used. Table \ref{tab:ablation} illustrates that removing the auto-thresholding and replacing it with an oracle-fixed threshold consistently degrades performance. When both the joint decision mechanism and auto-thresholding are removed, performance drops significantly across all datasets. As for the FNO layer, it outperforms the CNN-only architecture on HAR and EDF and achieves the second-best performance on HHAR. Conversely, removing both FNO and the joint decision mechanism results in the lowest overall scores. Overall, the joint decision mechanism consistently improves performance across all three datasets. These results demonstrate the contributions of this work, in particular the proposed joint decision mechanism and auto-thresholding for improving unknown class detection, as well as the benefits of using a TS–oriented architecture such as FNO.

\begin{table}[t]
\centering
\caption{Ablation study (H-scores)}
\renewcommand{\arraystretch}{1.25} 
\setlength{\tabcolsep}{5pt} 
\resizebox{0.9\linewidth}{!}{
\begin{tabular}{@{}cccccc@{}}
\toprule
\multicolumn{3}{c}{\textbf{Ablation}} & \multicolumn{3}{c}{\textbf{Datasets}} \\ 
\cmidrule(lr){1-3} \cmidrule(lr){4-6}
\textbf{Auto-Thresh.} & \textbf{Joint Decision} & \textbf{FNO} & \textbf{HAR} & \textbf{HHAR} & \textbf{EDF} \\ 
\midrule
\cmark                & \cmark                  & \cmark     & \textbf{62}  & \underline{53}  & \textbf{51}  \\ 
\cmark                & \cmark                  & \xmark     & 52           & \textbf{59}     & 47           \\ 
\cmark                & \xmark                  & \cmark     & 53           & 40              & 45           \\ 
\cmark                & \xmark                  & \xmark     & 39           & 43              & 42           \\ 
\xmark                & \cmark                  & \cmark     & \underline{61}  & 41           & \underline{49}  \\ 
\xmark                & \cmark                  & \xmark     & 19           & 39              & 43           \\ 
\xmark                & \xmark                  & \cmark     & 2            & 4               & 14           \\ 
\xmark                & \xmark                  & \xmark     & 1            & 5               & 12           \\ 
\bottomrule
\end{tabular}
}
\label{tab:ablation}
\end{table}

\section{Conclusion} \label{V}

In this work, we introduced UniJDOT, a novel OT-based method for UniDA, which outperformed all baseline methods on multiple TS datasets. The experiments showed that the output space of the model does not provide sufficient uncertainty on unknown target samples to accurately detect them. UniJDOT addresses this challenge by introducing a joint distance-regularized decision space, which improves uncertainty estimation. In addition, we have shown that the threshold is critical to performance and should be task-specific. This challenge is mitigated by our approach, which uses an auto-thresholding method from the image processing literature. Ablation studies confirmed that both proposed components consistently improve performance and ensure robustness without requiring extensive threshold fine-tuning. Finally, TS-oriented architectures were found to be effective on 2 of the 3 datasets, highlighting their importance. These results emphasize the need for further research on time-specific deep learning architectures for feature representation.

\section*{Acknowledgment}

The authors acknowledge the support of the French Agence Nationale de la Recherche (ANR), under grant ANR-23-CE23-0004 (project ODD) and of the STIC AmSud project DD-AnDet under grant N°51743NC.

\bibliography{main}
\bibliographystyle{ieeetr}
\appendix

\end{document}